\documentclass[10pt,twocolumn,letterpaper]{article}

\usepackage{iccv}
\usepackage{times}
\usepackage{epsfig}
\usepackage{graphicx}
\usepackage{amsmath}
\usepackage{amssymb}
\usepackage{multirow}
\usepackage{soul,xcolor}
\usepackage[accsupp]{axessibility}  

\usepackage{hyperref}
\hypersetup{pagebackref=true,breaklinks=true,letterpaper=true,colorlinks,bookmarks=false}

\iccvfinalcopy 

\graphicspath{{images/}}

\newcommand{\argmin}{\operatornamewithlimits{argmin}}

\def\x{{\bf x}}
\def\h{{\bf h}}

\def\z{{\bf z}}

\setstcolor{red}

\setstcolor{red}

\def\iccvPaperID{****} 

\ificcvfinal\fi

\pdfoutput=1

\begin{document}

\title{ScatSimCLR: self-supervised contrastive learning with pretext task regularization for small-scale datasets}

\author{Vitaliy Kinakh \,\,\,  Olga Taran \,\,\, Svyatoslav Voloshynovskiy\thanks{S. Voloshynovskiy is a corresponding author.}\\
Department of Computer Science, University of Geneva, Switzerland\\
{\tt\small \{vitaliy.kinakh, olga.taran, svolos\}@unige.ch}


}

\maketitle
\ificcvfinal\thispagestyle{empty}\fi

\begin{abstract}

 
 
In this paper, we consider a problem of self-supervised learning for small-scale datasets based on contrastive loss between multiple views of the data, which demonstrates the state-of-the-art performance in classification task. Despite the reported results, such factors as the complexity of training requiring complex architectures,  the needed number of views produced by data augmentation, and their impact on the  classification accuracy are understudied problems. To establish the role of these factors, we consider an architecture of contrastive loss system such as SimCLR, where baseline model is replaced by geometrically invariant “hand-crafted” network ScatNet with small trainable adapter network and argue that the number of parameters of the whole system and the number of views can be considerably reduced while practically preserving the same classification accuracy.
In addition, we investigate the impact of regularization strategies using pretext task learning based on an estimation of parameters of augmentation transform such as rotation and jigsaw permutation for both traditional baseline models and ScatNet based models.
Finally, we demonstrate that the proposed architecture with pretext task learning regularization achieves the state-of-the-art classification performance with a smaller number of trainable parameters and with reduced number of views.
Code: \href{https://github.com/vkinakh/scatsimclr}{https://github.com/vkinakh/scatsimclr}
\end{abstract}

\section{Introduction}
 
Self-supervised learning refers to the learning of data representations that are not based on labeled data. The recent techniques of self-supervised learning such as SimCLR \cite{chen2020simple},  SwAV \cite{caron2020unsupervised}, SeLa \cite{asano2019self} and BYOL \cite{grill2020bootstrap} demonstrate a classification performance close to their supervised counterparts. The main common idea behind these self-supervised approaches is to learn an embedding that produces an invariant representation under various data augmentations ranging from image filtering to geometrical transformations. In most cases, some powerful neural network such as for example ResNet \cite{chen2020simple} is used to implement this embedding. It is demonstrated \cite{chen2020simple} that the classification accuracy of these systems increases with the increase of the complexity of ResNet represented by the larger number of parameters capable of producing the invariance of visual representation under the broad family of augmentations. Typically the number of parameters of such networks ranges from 5M to 500M that makes their training quite a complex and time consuming task and requires a lot of training data. 

 \begin{figure}[t]
\centering
    \includegraphics[width=1\linewidth]{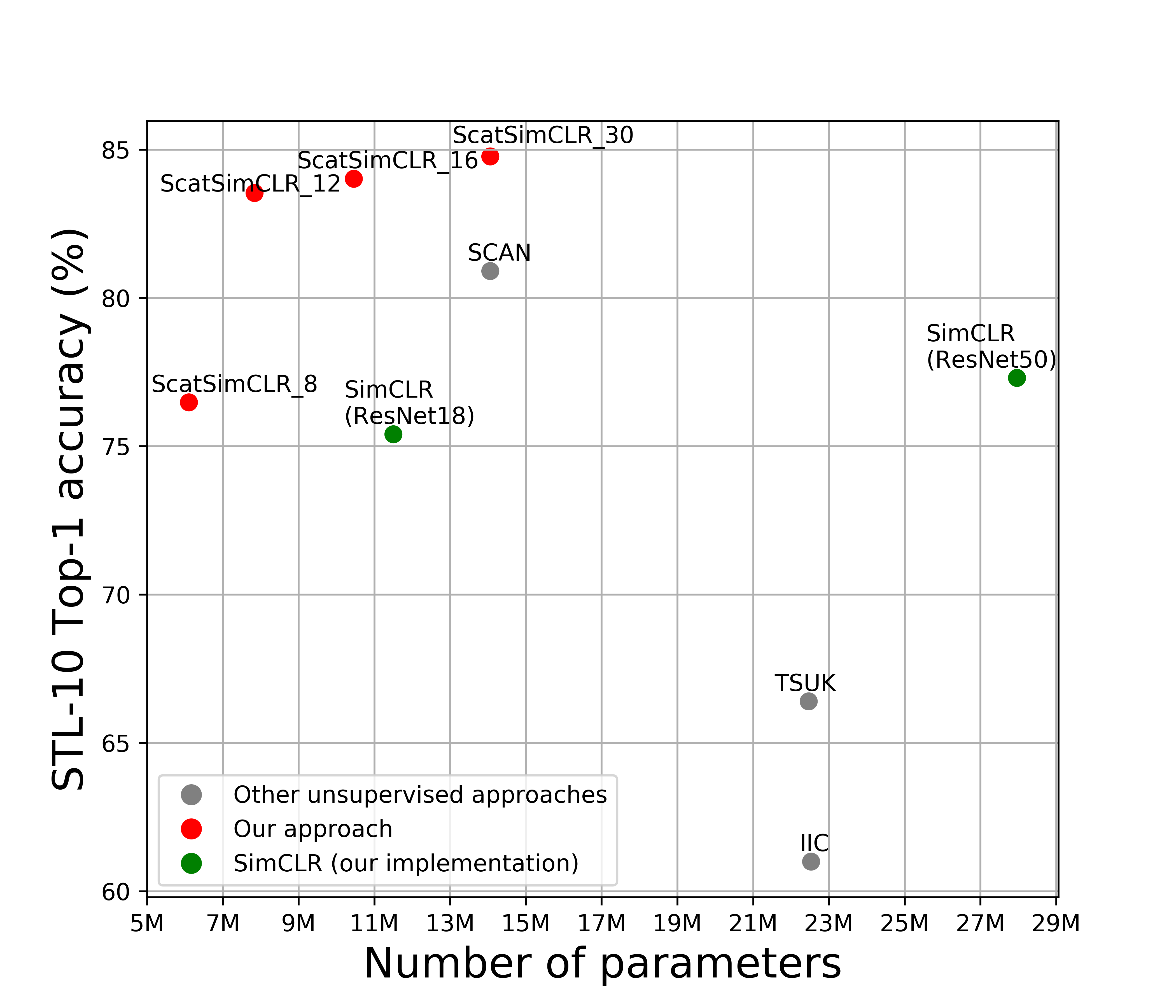}
    \caption{STL-10 \cite{pmlrv15coates11a} Top-1 accuracy of self-supervised methods. 
    Gray dots indicate other self-supervised methods. Our method, ScatSimCLR, is shown in red. Our implementation of SimCLR is shown in green. The results are obtained with models trained for 1000 epochs.}
\label{fig:perfromance1}
\end{figure}

At the time, in many practical applications it is infeasible to collect a lot of training data. Moreover, in many cases the amount of labeled data is limited. We refer to these cases as a "small dataset" problem. These restrictions lead to the overfitting of large scale models such as ResNet and result in their poor generalization. Therefore, to benefit from the recent advancements of self-supervised learning, which performance is generally demonstrated on the large scale datasets such as ImageNet \cite{deng2009imagenet}, it is important to develop efficient representation learning techniques adapted to the small dataset problem. 

In this paper, we try to address the problem of self-supervised learning based on contrastive loss in the application to the small dataset problem by replacing complex ResNet network by networks with a smaller number of parameters. More particularly, we investigate a question whether such complex networks as ResNet are really needed to achieve the targeted representation invariance assuming that the invariance to some families of augmentations can be achieved by a hand-crafted embedding. One candidate for such an invariant hand-crafted embedding is ScatNet \cite{andreux2020kymatio,bruna2013invariant}, which is known to produce stable embeddings under the deformations in terms of Lipschitz continuity property. As a by-product of such an invariance, one might assume that the number of augmentations needed for the training of invariant embedding can be reduced accordingly. Finally, the overall complexity of training might also be lower. To investigate these questions, we propose a ScatSimCLR architecture where the complex ResNet is replaced by ScatNet followed by a simple adapter network. We demonstrate that ScatSimCLR with a reduced number of training parameters and a reduced number of used augmentations can achieve similar performance and in some cases even outperform SimCLR. Furthermore, we demonstrate that the introduction of pretext task learning regularization, yet another popular technique of self-supervised learning, is beneficial for representation learning both for basic neural networks like SimCLR as well as for the proposed architectures.
 
{\bf Main contributions are:}
  \begin{enumerate}
	\item We propose a model with the reduced number of parameters of the embedding network while preserving the same classification performance. This is achieved due to the usage of the geometrically invariant network ScatNet. Figures \ref {fig:perfromance1} and \ref{fig:perfromance_cifar100} demonstrate the performance of ScatSimCLR on STL-10 and CIFAR100-20  \footnote{CIFAR100-20 is CIFAR100 dataset with 20 superclasses.} \cite{pmlrv15coates11a} as a function of the number of parameters with respect to the other state-of-the-art methods. The ScatSimCLR outperforms the state-of-the-art SCAN \cite{van2020scan} and RUC \cite{park2020improving} methods known to produce the top result for STL-10 and CIFAR100-20 datasets, while using even lower complexity networks.
	
	\item We investigate the impact of pretext task regularization on the classification performance. This includes the regularization based on the estimation of parameters of applied augmentation transform such as the rotation angle and jigsaw permutation.
	
\item We investigate the impact of the ScatNet and pretext task regularization on several datasets such as STL-10 and CIFAR100-20 and establish that the ScatSimCLR achieves state-of-the-art performance even with the smaller number of parameters.

 \item We investigate the role of augmentations in the context of representation learning based on the geometrically invariant ScatNet.

\item We demonstrate that the data agnostic ScatNet is applicable to the datasets with different statistics and labels and does not require extensive training as in the case of ResNet used for SimCLR contrastive learning.

\item Finally, we demonstrate that individual contributions of ScatNet and pretext tasks improves the performance of the model on classification tasks.
	
\end{enumerate}

\begin{figure}[t]
\centering
    \includegraphics[width=0.9\linewidth]{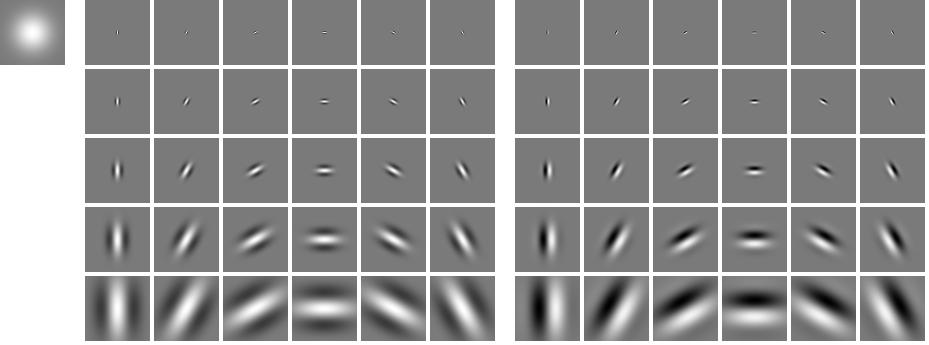}
    \caption{ScatNet \cite{andreux2020kymatio, bruna2013invariant} filter bank for J = 5 (number of scales) and L = 6 (number of rotations). The top left image corresponds to a low-pass filter.  The first left half image corresponds to the real parts of ScatNet filters arranged according to the scales (rows) and orientations (columns). The right half image corresponds to the imaginary part of ScatNet filters. }
\label{fig:scatnet_basis_functions}
\end{figure}

\section{Related work}
\label{sec:related_work}
 
We briefly summarize the related work to the concepts used in this paper.
 
 {\em Contrastive learning} is considered among the state-of-the-art techniques for self-supervised learning \cite{oord2018representation, hjelm2018learning, wu2018unsupervised, tian2019contrastive, sohn2016improved, chen2020simple}. The contrastive learning is based on a parameterized encoding or embedding that produces a low-dimensional data representation such that minimizes some distance between similar (positive) data pairs and maximizes for dissimilar (negative) ones. One of the central questions in contrastive learning is the generation or selection of positive and negative examples without labels. It is a common practice to generate positive examples by a data augmentation when multiple "views" for a given image are created by applying different crops \cite{pathak2016context, wu2018unsupervised, chen2020simple, bachman2019learning, he2020momentum, ye2019unsupervised, laskin2020curl}, various geometrical transformations of affine or projective families  \cite{gidaris2018unsupervised}, jigsaw image permutations \cite{noroozi2016unsupervised, doersch2015unsupervised}, splitting image into luminance and chrominance components \cite{tian2019contrastive}, applying low-pass and high-pass filtering \cite{chen2020simple, grill2020bootstrap}, predicting one view from another \cite{zhang2017split}, etc. The overall idea is to create a sort of "associations" between different parts of the same object or scene via a common latent space representation. The negative pairs are generally considered as images or parts of images randomly sampled from unlabeled data. The recent study \cite{tian2020makes} demonstrates the role of positive and negative example selection and generation and their impact on the overall classification accuracy.
 
\begin{figure}[t]
\centering
    \includegraphics[width=\linewidth]{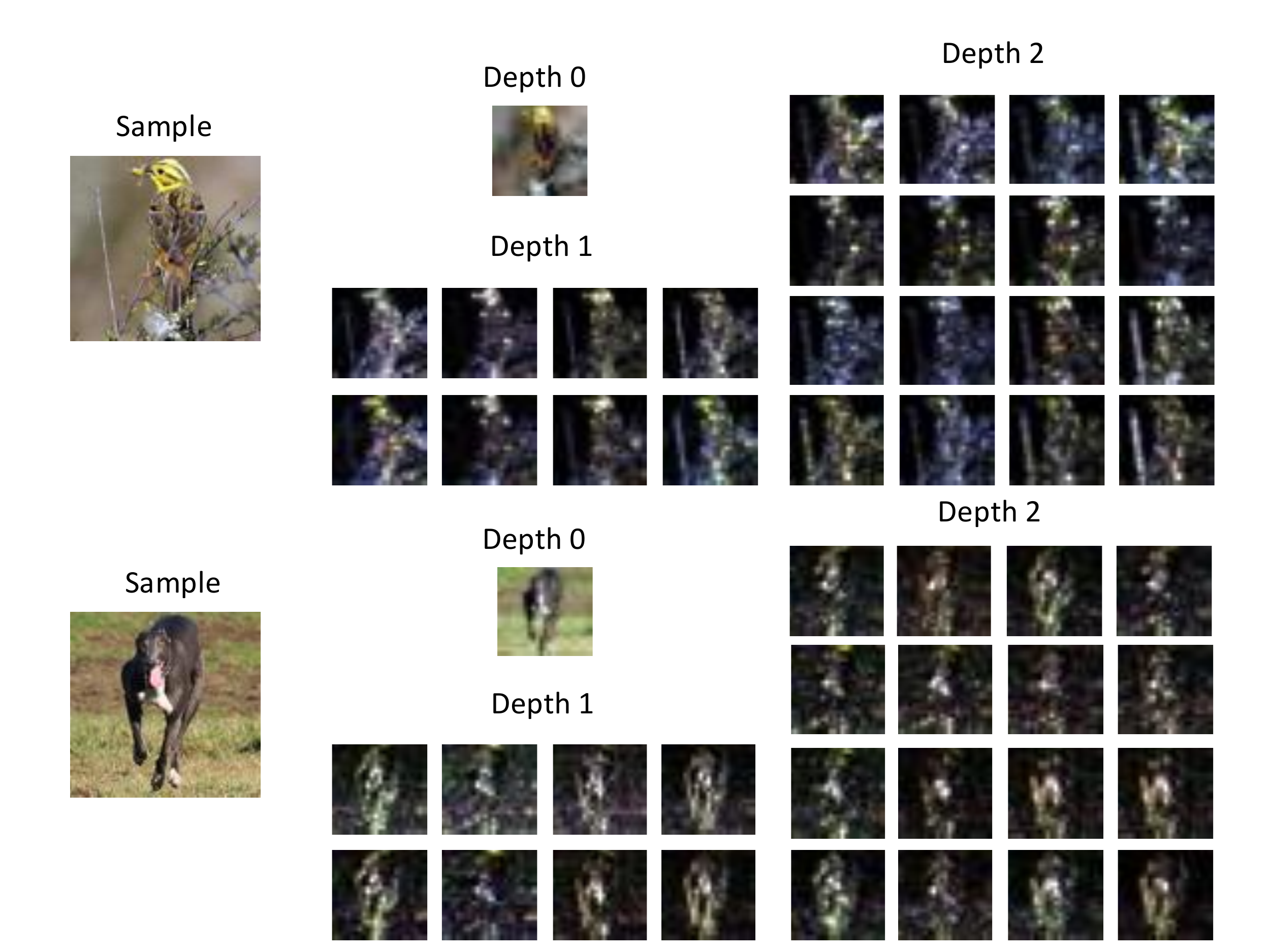}
    \caption{Examples of ScatNet \cite{andreux2020kymatio} feature vectors for $L=4$ and $J=2$ for STL-10 \cite{pmlrv15coates11a} images. ScatNet transform is applied to each color channel separately, then each channel is normalized and merged into a RGB image for better visualization.}
\label{fig:scatnet_example}
\end{figure}
 
\textit{Hand-crafted geometrically invariant transform - ScatNet\footnote{The efficient GPUs' implementation are provided in \cite{oyallon2018scattering, andreux2020kymatio}}} is a class of Convolutional Neural Networks (CNNs) designed with fixed weights \cite{bruna2013invariant} that has a set of important properties. (1) \textit{Deformation stability}: in contrast to the Fourier transformation that is generally unstable to  deformations at high frequencies\footnote{The Fourier transform is invariant to translation.}, ScatNet is stable to deformations in terms of Lipschitz continuity property. The stability is gained due to the use of non-linearity and average pooling. (2) \textit{Hand-crafted design}: ScatNet is considered as a deep convolution network with fixed filters in a form of wavelet basis functions independent of a specific dataset that at the same time provides (3) \textit{sparse representation}. (4) \textit{Interpretable representation}: in contrast to the most deep convolutional networks that output only the last layer, ScatNet outputs all layers representing the different signal scales. Figure \ref{fig:scatnet_basis_functions} shows typical ScatNet filters for the depth $J=5$ and number of orientations $L=6$. A set of features produced by ScatNet for the STL-10 \cite{pmlrv15coates11a} samples is shown in Figure \ref{fig:scatnet_example}.  

\textit{Hand-crafted pretext task and clustering based pseudo-labeling} are used to compensate for the lack of labeled data. The hand-crafted pretext task is considered as a sort of self-supervised learning when the input data are manipulated to extract a supervised signal in the form of a pretext task learning. 
The hand-crafted pretext task has been widely used in various settings to  predict the patch context \cite{doersch2015unsupervised, nathan2018improvements}, solve jigsaw puzzles from the same  \cite{noroozi2016unsupervised} and different images \cite{noroozi2018boosting}, colorize images \cite{zhang2016colorful, larsson2017colorization}, predict noise \cite{bojanowski2017unsupervised}, count \cite{noroozi2017representation}, estimate parameters of  rotations \cite{gidaris2018unsupervised}, inpaint patches  \cite{pathak2016context}, spot artifacts \cite{jenni2018self}, generate images \cite{ren2018cross} as well as for  predictive coding \cite{oord2018representation, henaff2019data} and instance discrimination \cite{wu2018unsupervised, he2020momentum, chen2020simple, tian2019contrastive, misra2020self}. We refer the reader to \cite{jing2020self} for the details of these methods. At the same time, clustering based pseudo-labeling can be used as pseudo-labels to learn visual representations  \cite{caron2018deep}. Recent work \cite{caron2020unsupervised} extends this idea to soft cluster assignment in contrast to hard-assignment. In this work we only consider pretext task learning based on rotation and jigsaw parameters' estimation.


\section{ScatSimCLR}

\begin{figure}[t]
\centering
    \includegraphics[width=1\linewidth]{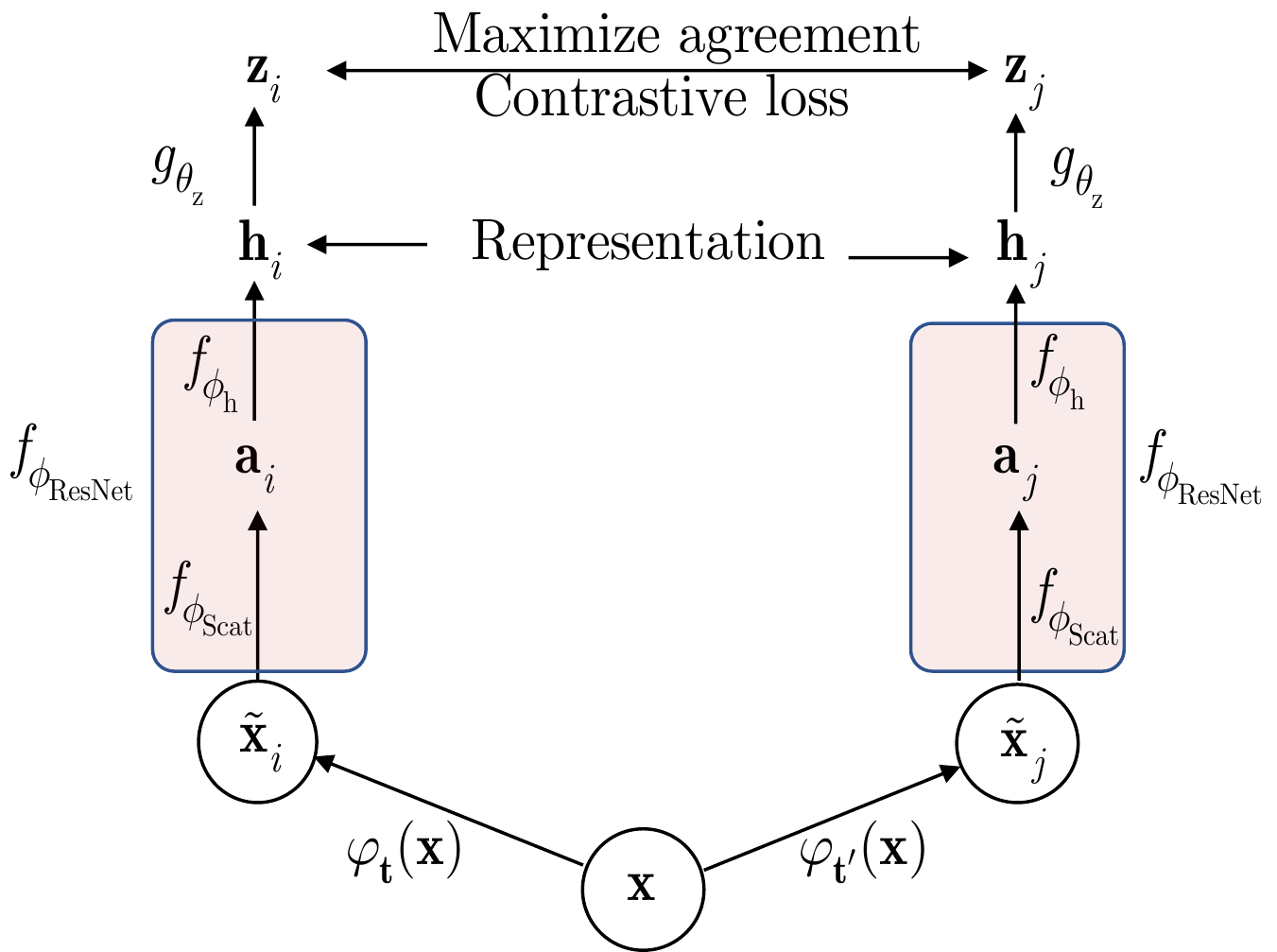}
    \caption{Contrastive learning of visual representation according to SimCLR architecture. In this work, an encoding network $f_{{\boldsymbol \phi}_{\mathrm{ResNet}}}$ producing a representation $\bf h$ is replaced by ScatNet network $f_{{\boldsymbol \phi}_{\mathrm{Scat}}}$ and adapter network  $f_{{\boldsymbol \phi}_{\mathrm{h}}}$. In the rest, the architecture remains the same as for SimCLR. }
\label{fig:SimCLR}
\end{figure}

The proposed architecture of self-supervised representation learning is shown in Figure \ref{fig:SimCLR} and it is based on the SimCLR framework.

For the batch size $N$, given $\{\x_k\}_{k=1}^{N}$ in the batch, SimCLR produces two augmented versions $\tilde{\x}_{2 k-1}=\varphi_\mathrm{\bf t} \left(\x_{k}\right)$ and $\tilde{\x}_{2 k}=\varphi_\mathrm{\bf t'} \left(\x_{k}\right)$ of each $\x_k$ using parameterized augmentation transform $\varphi_\mathrm{\bf t}$ with parameters $\bf t$ and $\bf t'$ for each view. Both augmented images are first processed by the feature extraction network $f_{{\boldsymbol \phi}_{\mathrm{ResNet}}}$ thus producing two representations $\boldsymbol{\h}_{2 k-1}=f_{{\boldsymbol \phi}_{\mathrm{ResNet}}}\left(\tilde{\boldsymbol{\x}}_{2 k-1}\right)$ and $\boldsymbol{\h}_{2 k}=f_{{\boldsymbol \phi}_{\mathrm{ResNet}}}\left(\tilde{\boldsymbol{\x}}_{2 k}\right)$ and then by the projection network $g_{{\boldsymbol \theta}_{\mathrm{z}}}$ that produces two vectors $\boldsymbol{\z}_{2 k-1}=g_{{\boldsymbol \theta}_{\mathrm{z}}}
\left(\boldsymbol{\h}_{2 k-1}\right)$ and $\boldsymbol{\z}_{2 k}=g_{{\boldsymbol \theta}_{\mathrm{z}}}
\left(\boldsymbol{\h}_{2 k}\right)$.

SimCLR contrastive loss is defined as: 

\begin{equation}
\begin{aligned}
\label{eq:SimCLR_loss}
  \mathcal{L}_{\mathrm{SimCLR}} ({\boldsymbol \phi}_{\mathrm{ResNet}},{\boldsymbol \theta}_{\mathrm{z}}) = \frac{1}{2 N} \sum_{k=1}^{N}[\ell(2 k-1,2 k)+ \\ \ell(2 k, 2 k-1)],
\end{aligned}
\end{equation}

 where $\ell(i, j)=-\log \frac{\exp \left(s_{i, j} / \tau\right)}{\sum_{k=1}^{2 N} 1_{\mid k \neq i]} \exp \left(s_{i, k} / \tau\right)}$
  with $s_{i, j}={\z}_{i}^{\top} {\z}_{j} /\left(\left\|{\z}_{i}\right\|\left\|{\z}_{j}\right\|\right) $ denotes a pairwise similarity for all pairs $i \in\{1, \ldots, 2 N\}$ and $j \in\{1, \ldots, 2 N\}$ and $1_{[k \neq i]} \in\{0,1\}$  is an indicator function evaluating to
1 iff $k \neq i$ and $\tau$ denotes a temperature parameter.

SimCLR demonstrates the increasing performance in classification accuracy as shown in Figure \ref{fig:perfromance1} with the growth of the number of parameters of ResNet $f_{{\boldsymbol \phi}_{\mathrm{ResNet}}}$ from about 11M to 28M. Thus it is commonly assumed that this increase in performance is due to the increase of $f_{{\boldsymbol \phi}_{\mathrm{ResNet}}}$  network capacity and its ability to learn more complex associations between different parts of objects. Obviously, all the parameters of the network should be trained to efficiently encode these associations.

In contrast to this, we argue that the complex trainable ResNet $f_{{\boldsymbol \phi}_{\mathrm{ResNet}}}$ can be replaced by the hand-crafted non-trainable ScatNet network $f_{{\boldsymbol \phi}_{\mathrm{Scat}}}$ and small capacity trainable adapter network $f_{{\boldsymbol \phi}_{\mathrm{h}}}$. ScatNet network $f_{{\boldsymbol \phi}_{\mathrm{Scat}}}$ is a hand-crafted network with the fixed parameters and it is agnostic to a particular dataset and corresponding inter-object associations. It produces invariant low-level image representation $\bf a$. At the same time, the low capacity adapter network $f_{{\boldsymbol \phi}_{\mathrm{h}}}$ aggregates the output of ScatNet and produces the visual representation $\bf h$. Therefore, one should only train the parameters of an adapter network that is just a fraction of ResNet. Similarly to the results presented in  Figure \ref{fig:perfromance1}, one can change the complexity of the adapter network and investigate its impact on the overall system performance. For the fair comparison, we keep the remaining architecture the same as in SimCLR.

\begin{figure}[t]
\centering
    \includegraphics[width=0.7\linewidth]{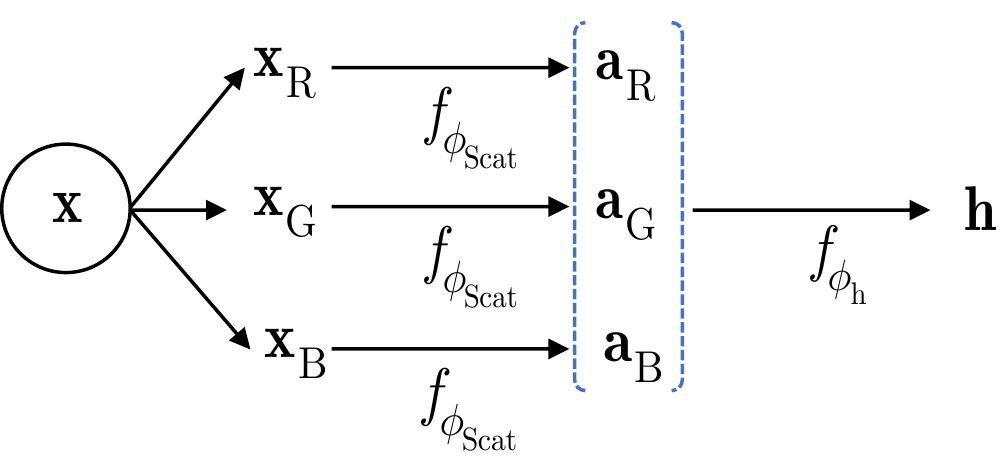}
    \caption{The encoding network for color images. An image $\bf x$ is represented by three color components $\{ {\bf x}_{\mathrm{R}}, {\bf x}_{\mathrm{G}}, {\bf x}_{\mathrm{B}}\}$.  Each color component is processed by ScatNet network $f_{{\boldsymbol \phi}_{\mathrm{Scat}}}$ and then the adapter network $f_{{\boldsymbol \phi}_{\mathrm{h}}}$ aggregates the outputs to produce the representation $\bf h$. }
\label{fig:color_Scatntesimclar}
\end{figure}

To process color images, we simply apply  ScatNet to each color channel as shown in Figure \ref{fig:color_Scatntesimclar}. We have used RGB representation but YCbCr or YUV spaces might be even more suited due to the properties of Y component reflecting grayscale images.

We will refer to SimCLR network with the replaced ResNet by ScatNet and the adapter network as ScatSimCLR.

\section{Additional regularizer as a pretext task self-learning} 
\label{sec:ScatSimCLR_reg}
 
 In this section, we introduce an additional form of regularization that does not require any labeling, pseudo-labeling or mining for positive or negative neighbors as in \cite{ van2020scan} that can be surely applied to our framework. Instead for fair comparison with SimCLR we will stay in the scope of the same self-supervised framework and try to explore another direction by investigating the role of latent space regularization via {\em estimation of parameters of applied augmentation transformation}. The pretext task regularization methods are not new and have been used in a stand-alone self-supervised architectures as described in Section 2.  However, up to our best knowledge these regularization techniques have not been considered in the scope of contrastive representation learning. Thus a hypothesis to verify is whether creating more semantics about the inter-object or inter-scene associations would lead to more meaningful latent space representation.
 
\begin{figure}[t]
\centering
    \includegraphics[width=1\linewidth]{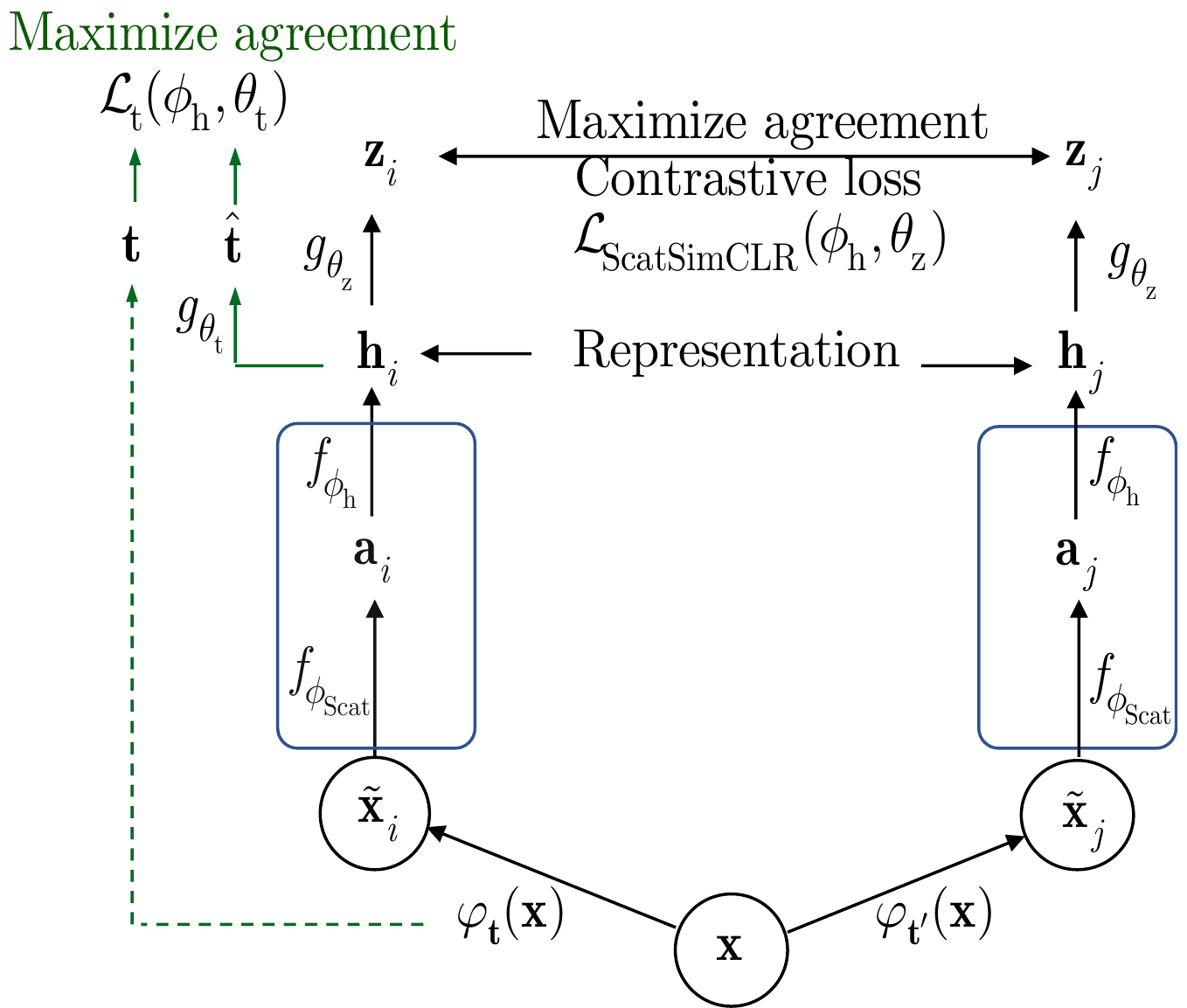}
    \caption{ScatSimCLR with an additional regularization based on the estimation of parameter $\bf t$ and $\bf t'$ of augmentation transform $\varphi_t$ via a network $g_{{\boldsymbol \theta}_{\mathrm{t}}}$ applied to both left and right channels (schematically shown only for the left channel in green).
    }
\label{fig:ScatSimCLR}
\end{figure}
 
 In our study we define the parameters $\mathrm{\bf t}$ of the augmentation transform ${\varphi}_{\mathrm{\bf t}}$ under the pretext task estimation to be the rotation or jigsaw permutation. We used 4 rotation angles (0\textdegree, 90\textdegree, 180\textdegree and 270\textdegree) and 35 jigsaw permutations. We apply only one augmentation (either rotation or jigsaw permutation) at time. These parameters are encoded as one-hot-encoding for each augmentation and the corresponding classifier $g_{{\boldsymbol \theta}_{\mathrm{\bf t}}}$ is used to estimate them from the visual representation ${\bf h} = f_{{\boldsymbol \phi}_{\mathrm{Scat}}} ( f_{{\boldsymbol \phi}_{\mathrm{h}}}({\bf \tilde x}) )$ extracted from the augmented view ${\bf \tilde x} = {\varphi}_{\mathrm{t}}({\bf \tilde x})$ as shown in Figure \ref{fig:ScatSimCLR}.

 The pretext task loss is defined as the parameters estimation loss between the applied parameters $\mathrm{\bf t}$ and estimated ones $ \hat{\mathrm{\bf t} } = g_{{\boldsymbol \theta}_{\mathrm{t}}}({\bf h})$:
 
 \begin{equation} 
\label{eq:pretext task_loss}
  \mathcal{L}_{\mathrm{t}} ({\boldsymbol \phi}_{\mathrm{h}}, {\boldsymbol \theta}_{\mathrm{t}} ) = \frac{1}{2N}\sum_{i=1}^{2N} d(\mathrm{ \bf t}_i, g_{{\boldsymbol \theta}_{\mathrm{t}}}({\bf h}_i)), 
\end{equation}
where $d(.,.)$ denotes the cross-entropy.

\section{Final loss and training}
 \label{Loss}
 
 We define ScatSimCLR loss similarly to SimCLR loss (\ref{eq:SimCLR_loss}) with the only difference that instead of ${\bf h} = f_{{\boldsymbol \phi}_{\mathrm{ResNet}}}({\bf \tilde x})$, we consider ${\bf h} = f_{{\boldsymbol \phi}_{\mathrm{h}}} ( f_{{\boldsymbol \phi}_{\mathrm{Scat}}}({\bf \tilde x}) )$. Thus, the loss of ScatSimCLR is denoted as  $  \mathcal{L}_{\mathrm{ScatSimCLR}} ({\boldsymbol \phi}_{\mathrm{h}},{\boldsymbol \theta}_{\mathrm{z}})$.
 
 The final loss of ScatSimCLR with the pretext task regularization is defined as:
 
\begin{equation} 
\label{eq:final_loss}
 \mathcal{L}({\boldsymbol \phi}_{\mathrm{h}},{\boldsymbol \theta}_{\mathrm{z}}, {\boldsymbol \theta}_{\mathrm{t}} ) = \mathcal{L}_{\mathrm{ScatSimCLR}} ({\boldsymbol \phi}_{\mathrm{h}},{\boldsymbol \theta}_{\mathrm{z}}) + \lambda \mathcal{L}_{\mathrm{t}} ({\boldsymbol \phi}_{\mathrm{h}}, {\boldsymbol \theta}_{\mathrm{t}} ),
\end{equation}
where $\lambda$ controls the relative weight of the second loss term.

The parameters estimation is based on the minimization problem:
\begin{equation} 
\label{eq:final_loss_minimization}
 ( \hat{\boldsymbol \phi}_{\mathrm{h}}, \hat{\boldsymbol \theta}_{\mathrm{z}},  \hat{\boldsymbol \theta}_{\mathrm{t}})  = \argmin_{({\boldsymbol \phi}_{\mathrm{h}},{\boldsymbol \theta}_{\mathrm{z}}, {\boldsymbol \theta}_{\mathrm{t}})}
 \mathcal{L}({\boldsymbol \phi}_{\mathrm{h}},{\boldsymbol \theta}_{\mathrm{z}}, {\boldsymbol \theta}_{\mathrm{t}} ), 
\end{equation}
in practical implementation for the first 40 epochs we assume $\lambda = 0$ in (\ref{eq:final_loss}) and then $\lambda = 0.3$ for the rest. We have noticed that the network converges better, if it is pre-trained with only contrastive loss at the beginning. The parameter $\lambda$ is selected to equalize the amplitude of contrastive and cross-entropy losses. 

 \section{Experimental results}
  \label{sec:results}

In this section, we evaluate ScatSimCLR performance on several datasets in the image classification task. At first, the proposed model is pretrained on a particular dataset based on (\ref{eq:final_loss_minimization}) using unlabeled data and then a logistic one-layer classifier is applied to the learned representation to map it to the class labels encoded based on one-hot-vector encoding.  

\textbf{Datasets.} The experimental evaluation is performed on STL-10 \cite{pmlrv15coates11a} and CIFAR100-20 \cite{krizhevsky09learningmultiple} datasets. The experiments aim at investigating the impact of ScatSimCLR architecture and image augmentations on the classification performance. The results are reported as a top-1 result from 5 different runs.

\subsection{Impact of ScatSimCLR parameters}

\subsubsection{Impact of scaling and rotation channels}
\label{lab:scale_rotation}

\begin{table}[]
\caption{Impact of scale J and rotations L parameters of ScatNet on the classification accuracy after 5 epochs. 
\label{table:scatnet_params}}

\begin{tabular}{llcc}
\hline
J & L  & Accuracy STL-10 & Accuracy CIFAR100-20 \\ \hline
1 & 4  & 61.90\%         & 36.88\%           \\ 
1 & 8  & 62.75\%         & 39.43\%           \\
1 & 12 & 63.00\%         & 40.56\%           \\
1 & 16 & 63.70\%         & 41.72\%           \\ \hline
2 & 4  & 63.12\%         & 44.52\%           \\
2 & 8  & 63.71\%         & 46.09\%           \\
2 & 12 & 63.34\%         & 46.25\%           \\
2 & 16 & \textbf{64.03\%}         & \textbf{46.73\%}           \\ \hline
3 & 4  & 60.10\%         & 42.50\%           \\
3 & 8  & 60.80\%         & 43.85\%           \\
3 & 12 & 60.90\%         & 44.01\%           \\
3 & 16 & 61.20\%         & 44.59\%           \\ \hline
4 & 4  & 45.12\%         & 34.39\%           \\
4 & 8  & 46.58\%         & 35.00\%           \\
4 & 12 & 48.10\%         & 35.96\%           \\
4 & 16 & 49.91\%         & 36.74\%           \\ \hline
\end{tabular}
\end{table}




In this section, we investigate the impact of ScatNet parameters on the overall performance of ScatSimCLR. We use two datasets STL-10 and CIFAR100-20 with the images of  size 96x96 to fit ScatNet. It should be noted that CIFAR100-20 is up-sampled from the size 32x32 to 96x96 using LANCZOS interpolation \cite{Lanczos1950AnIM}. The system is trained with respect to the contrastive loss $\mathcal{L}_{\mathrm{ScatSimCLR}} ({\boldsymbol \phi}_{\mathrm{h}},{\boldsymbol \theta}_{\mathrm{z}})$ and with adapter network fixed to 12 ResBlock layers and fixed depth of ScatNet to be 2. The pretext task loss was not used and the training was performed for the first 5 epochs only to reflect the dynamics of learning.

We have considered a range of ScatNet scaling parameters $J$ from 1 to 4. We experimentally established that for the current architecture of ScatNet applied to the investigated datasets with the images of size 96x96, the best scaling parameter $J$ is 2 as shown in Table \ref{table:scatnet_params}. It should be pointed out that the increase of the scaling leads to the usage of larger filter sizes. As a consequence, the size of resulting images on the output of ScatNet, representing the feature vector, decreases. In turns, this represents a trade-off between the desirable robustness to the scaling and undesirable loss of details in the produced images. This might explain the optimality of the scaling factor $J$=2 as opposed to $J$=4.

Table \ref{table:scatnet_params} also demonstrates the impact of rotation parameter $L$ on the classification performance for the considered scale factors $J$. The investigation of the rotation parameter $L$ was performed in the range from 4 till 16 with the step size equals to 4 for each scale factor. For both considered datasets, the increase of the number of rotations clearly leads to the increase of the classification performance that can be explained by the increase of the rotation invariance in the produced feature space. In contrast to the scaling, the increase of the rotation factor $L$ preserves the dimensionality of the produced feature map for a given fixed scaling and only leads to the increase of the number of rotation channels in the network output. This might explain the increase of the rotation parameter leads to overall performance enhancement.

\subsubsection{Impact of the number of layers in the adapter network}
\label{lab:adapter}
In this section, we investigate the impact of the adapter network parameters on the classification accuracy. The experiments are performed on the dataset STL-10 with the image size 96x96. The training loss is defined by (\ref{eq:final_loss}). As the pretext task network we used a classifier consisting of two fully-connected layers followed by the traditional dropout and ReLu activation. The last layer activation is softmax. ScatNet parameters were chosen according to the best results of section \ref{lab:scale_rotation}, i.e., $J$=2 and $L$=16.

\begin{table}[]
\caption{Impact of the number of layers in the adapter network of ScatSimCLR on STL-10 dataset for 1000 epochs. \label{table:adapter_network}}
\scalebox{0.96}{%
\begin{tabular}{ccc}
\hline
\multicolumn{1}{l}{Num. of layers} & \multicolumn{1}{l}{Num. of parameters} & \multicolumn{1}{l}{Accuracy STL-10} \\ \hline
8                                    & 6.1M                                    & 76.47\%                              \\
12                                   & 7.8M                                    & 83.53\%                              \\
16                                   & 10.4M                                   & 84.01\%                              \\
30                                   & 14.1M                                   & 85.11\%                              \\ \hline
\end{tabular}}
\end{table}

We investigate the adapter network with the different number of layers, namely 8, 12, 16 and 30.
ScatSimCLR was trained during 1000 epochs for each considered adapter network. The results presented in Table \ref{table:adapter_network} are obtained as the top-1 results on the validation set. The obtained results clearly indicate that the increase of the adapter network complexity increases the performance in the classification task.

\subsection{Ablations}

\subsubsection{Regularization ablations}

In this section we investigate the impact of the regularization techniques. We compare the performance of the model trained with and without pretext task based on the estimation of augmentation transform: rotation and jigsaw estimation. We run experiment with all models presented in Table \ref{table:adapter_network} and ScatSimCLR based on ResNet18 to compare the performance in a function of model complexity. ScatNet parameters were chosen according to the best results presented in \ref{lab:scale_rotation}. We use the STL-10 dataset for our comparison experiments. As the pretext task estimator we used a classifier consisting of two fully-connected layers with ReLU activation and the softmax at the end. We trained each model for 100 epochs. The batch size differs depending on the size of the model.

As it is shown in Table \ref{table:regularization_impact}, the introduction of the pretext task improves the classification accuracy for both considered models: (i) ScatNet based SimCLR and (ii) vanilla SimCLR.
For all models, rotation augmentation pretext task provides higher increase in classification performance in comparison to jigsaw. It can be explained by the fact that in the process of jigsaw pretext task, an image is split into 9 patches without an intersection, and then each patch is resized using Lanczos interpolation, so they fit the network input size. Applying interpolation introduces some artifacts. In the considered pretext task based on rotating by 90, 180 and 270 degrees the interpolation is not applied as such.

Therefore the introduction of pretext task regularization improves the classification performance of the models trained with contrastive loss. The proposed ScatSimCLR8 with 6.1 M of parameters outperforms SimCLR (ResNet18) with 11.5 M of parameters for all considered pretext tasks and also without pretext task. This confirms the importance of geometrical invariance of ScatNet.




\begin{table}[]
\caption{Impact of the pretext task regularization on the classification accuracy on  STL-10 dataset.}
\label{table:regularization_impact}
\scalebox{0.85}{
{
\begin{tabular}{lcccc}
\multicolumn{5}{c}{Accuracy on STL-10} \\ \hline
\multicolumn{1}{c}{\multirow{2}{*}{\begin{tabular}[c]{@{}c@{}}Baseline\\  model\end{tabular}}} & \multirow{2}{*}{\begin{tabular}[c]{@{}c@{}}Without \\ pretext\end{tabular}} & \multicolumn{2}{c}{With pretext}            & \multirow{2}{*}{\begin{tabular}[c]{@{}c@{}}Num.\\ of paramers\end{tabular}} \\ \cline{3-4}
\multicolumn{1}{c}{} & & Rotation & Jigsaw & \\ \hline
ScatSimCLR 8      & 74.78\% & \textbf{77.86\%} & 76.36\% & 6.1 M  \\ \hline
ScatSimCLR 12     & 76.57\% & \textbf{78.43\%} & 77.78\% & 7.8 M  \\ \hline
ScatSimCLR 16     & 77.03\% & \textbf{78.5\%}  & 77.91\%    & 10.5 M \\ \hline
ScatSimCLR 30     & 77.86\% & \textbf{79.11\%} & 78.4\%    & 14.1 M  \\ \hline
SimCLR (ResNet18) & 71.90\% & \textbf{76.36\%} & 75.22\%   & 11.5 M  \\ \hline
\end{tabular}}}
\end{table}

\subsubsection{Ablations of image augmentations}

In this section we investigate the impact of image augmentations on the classification performance. We use the STL-10 dataset with image size 96x96. To exclude the impact of batch size and other model hyperparameters, we use the fixed setup with batch size = 256, ScatNet parameters: $J$=2, $L$=16 and depth=2. We study \textit{(i)} {\em geometric transformations}: random cropping, horizontal flipping and random affine transformations and \textit{(ii)} {\em color transformations}: color jitter, Gaussian blur and grayscaling. We tried to investigate the effect of augmentation ablation considering different combinatorics of augmentations.

\begin{figure}[t]
\centering
    \includegraphics[width=1\linewidth]{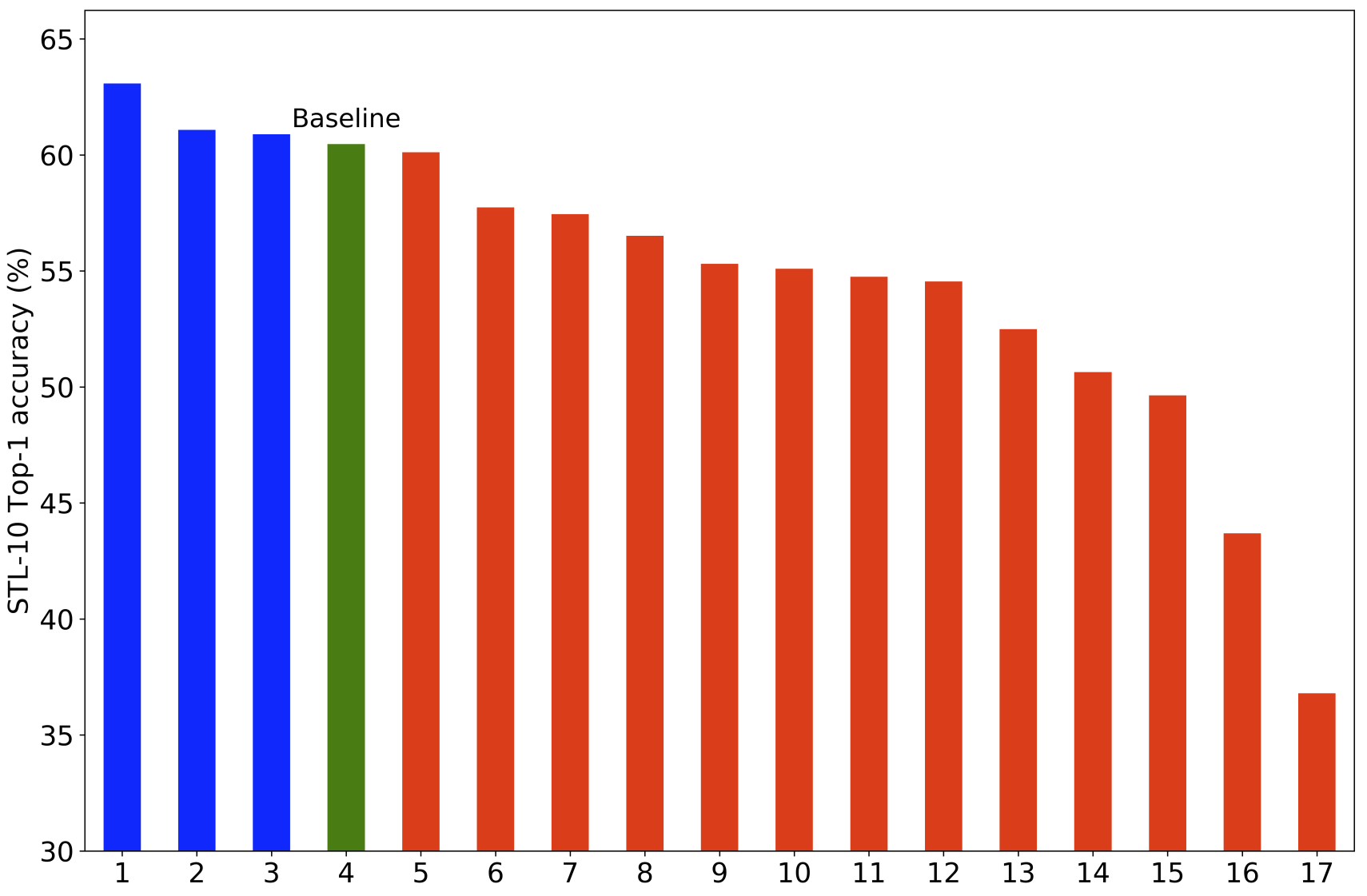}
    \caption{Impact of removing the augmentations on the performance of ScatSimCLR for STL-10: "Baseline" denotes ScatSimCLR trained with all augmentations (cropping, flipping, color, grayscale, Gaussian blur and affine augmentations). The following labels denote: 1 - the baseline without the affine augmentations; 2 - only cropping and color augmentations; 3 - the baseline without the horizontal flipping; 5 - the baseline without Gaussian blur augmentations; 6 - the baseline without cropping and Gaussian blur augmentations; 7 - the baseline without color and Gaussian blur augmentations; 8 - the baseline without grayscale and Gaussian blur augmentations; 9 - the baseline without cropping and grayscale augmentations; 10 - the baseline without color augmentations; 11 - the baseline without cropping augmentations; 12 - the baseline without grayscale augmentations; 13 - only cropping augmentations; 14 - the baseline without color and grayscale augmentations; 15 - only color augmentations; 16 - the baseline without crop and color augmentations; 17 - no augmentations.}
\label{fig:impact_augmentations}
\end{figure}

The obtained results are shown in Figure \ref{fig:impact_augmentations}. The baseline system performance is shown by the green bar. The baseline uses all considered augmentation similarly to SimCLR. It is interesting to point out that the removal of affine transformation augmentations leads to the performance enhancement with about 2\% with respect to the baseline system with all considered augmentations. This is an important result confirming the invariance of ScatNet to geometrical transformations. Therefore, these augmentations can be further excluded from training. In turns, it might lead to the lower complexity of training under a smaller number of augmentations. The next interesting result is obtained when the only image cropping and color transformations were used as the augmentations. It leads to about 0.5\% enhancement over the baseline system. Finally, the same enhancement is observed when the flipping was removed from the baseline augmentations. The performance of ScatSimCLR without any augmentations is about 24\% lower with respect to the baseline system.  

Summarizing the obtained results, we can conclude that the most important augmentations for ScatSimCLR are cropping and color ones. 




 \subsubsection{Comparison with the state-of-the-art}
 
\begin{figure}[t]
\centering
    \includegraphics[width=1\linewidth]{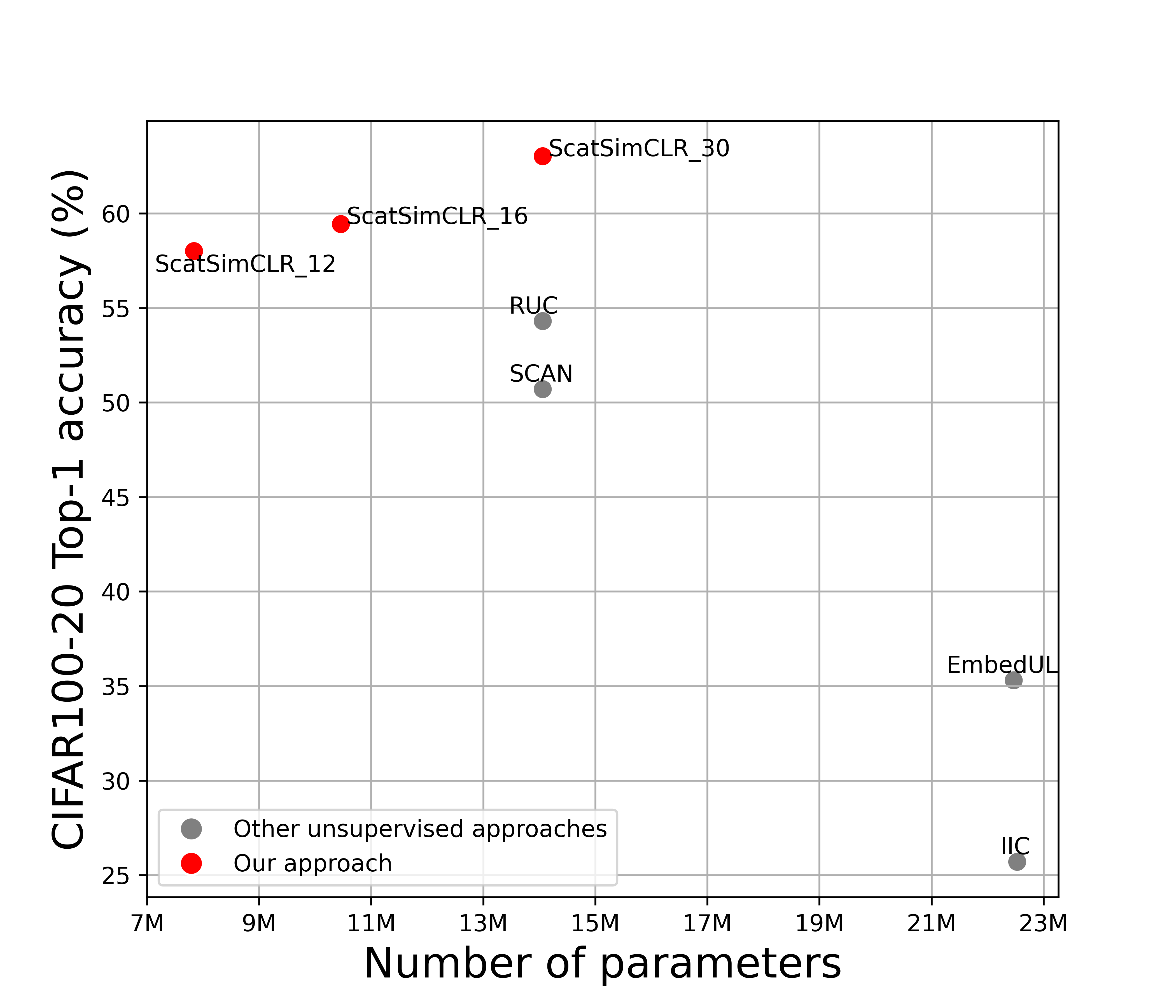}
    \caption{CIFAR100-20 Top-1 accuracy of self-supervised methods. 
    Gray dots indicate other self-supervised methods.  ScatSimCLR is shown in red.}
\label{fig:perfromance_cifar100}
\end{figure}

 We compare the results obtained for the proposed ScatSimCLR on STL-10  \cite{pmlrv15coates11a} and CIFAR100-20 \cite{krizhevsky09learningmultiple} with the state-of-the-art results reported in ADC \cite{haeusser2019adc}, DeepCluster \cite{caron2018deep}, DAC \cite{chang2017dac}, IIC \cite{ji2019invariant}, TSUK \cite{han2020mitigating}, SCAN \cite{van2020scan}, RUC \cite{park2020improving} and SimCLR\cite{chen2020simple} on the Figures \ref{fig:perfromance1} and \ref{fig:perfromance_cifar100}.


Figures \ref{fig:perfromance1} and \ref{fig:perfromance_cifar100} show the performance of image classification using ScatSimCLR with the linear evaluation layer. We compare the model performance not only in terms of classification accuracy but also in terms of number of trainable parameters. For the STL-10 dataset as shown on Figure \ref{fig:perfromance1}, we not only achieve SOTA classification accuracy but also our model achieves better performance, compared to previous SOTA \cite{van2020scan} with only a half of its parameters. The same tendency is shown for CIFAR100-20 \cite{krizhevsky09learningmultiple} dataset on Figure \ref{fig:perfromance_cifar100}; all proposed ScatSimCLR models achieve SOTA classification accuracy: 58.0\% , 59.4\% and 63.8\%, with 7M, 10.4M and 14M parameters respectively, while previous SOTA RUC \cite{park2020improving} achieves 54.3\% with 14M trainable parameters. 


 
 \section{Conclusion and discussions}
 
 
 In this paper,  we address the problem of self-supervised learning for small dataset problems. More particularly, we answer the question whether the complex encoding network used for the contrastive learning can be partially replaced by the simpler hand-crafted  network  ensuring geometric invariance.
 
 We demonstrate that the proposed model based on geometrically invariant ScatNet with reduced number of trainable parameters can achieve the state-of-the-art performance on STL-10 and CIFAR100-20 datasets. 
 
 We demonstrate that introduction of pretext task regularization based on the estimation of augmentation transform improves the performance of the proposed ScatSimCLR models as well as SimCLR with ResNet.
 
 We demonstrate that by using a geometrically invariant ScatNet model, we are able to reduce the great portion of augmentations used to simulate the geometrical transformations at the training. Also, we confirm that the main benefit in the considered contrastive learning comes  from the color and cropping  augmentations. This indicates that a promising direction in further reduction of the number of augmentations is to use more efficient color coding schemes and to introduce local windowed encoding in contrast to the whole image encoding considered in the paper. 

  The performed extensive experiments  explain  the  architectural  and  design  particularities of the considered approach.  The obtained results represent the state-of-the-art performance on several datasets among the networks with the same number of parameters.
 

\section*{Acknowledgement}
This research was partially funded by the Swiss National Science Foundation SNF project CRSII5\_193716.

{\small
\bibliographystyle{ieee_fullname}
\bibliography{egbib}
}

\end{document}